%% file: main.tex
\documentclass[10pt, a4paper]{article}

\usepackage[final]{lrec-coling2024}

\usepackage[absolute]{textpos}

\input{macros}

\title{OOVs in the Spotlight: How to Inflect them?}

\name{Tomáš Sourada, Jana Straková, Rudolf Rosa}
\address{Institute of Formal and Applied Linguistics, Faculty of Mathematics and Physics\\
Charles University, Czech Republic\\
\texttt{\{sourada,strakova,rosa\}@ufal.mff.cuni.cz}}

\abstract{\input{00_abstract_short}
 \\ \newline  }

\begin{document}

\begin{textblock}{16}(0,0.1)\centerline{This paper was published in \textbf{LREC-COLING 2024}}\end{textblock}
\begin{textblock}{16}(0,0.3)\centerline{-- please cite the published version {\small\url{https://aclanthology.org/2024.lrec-main.1091}}.}\end{textblock}

\maketitleabstract

\input{01_intro}

\input{02_rel}
\input{03_data}
\input{04_methods}
\input{05_eval}
\input{06_error}
\input{07_concl}
\input{08_limitations}
\input{09_ethical}
\input{10_ack}

\section{Bibliographical References}\label{sec:reference}

\bibliographystyle{lrec-coling2024-natbib}
\bibliography{lrec-coling2024}

\section{Language Resource References}
\label{lr:ref}
\bibliographystylelanguageresource{lrec-coling2024-natbib}
\bibliographylanguageresource{languageresource}

\end{document}

%% file: macros.tex
\usepackage[textsize=footnotesize, backgroundcolor=yellow!25, linecolor=black!25]{todonotes}

\usepackage{makecell}
\usepackage{multirow}
\usepackage{multicol}
\usepackage{amsmath}

\usetikzlibrary{positioning}
\usetikzlibrary{arrows.meta}
\usetikzlibrary{decorations.pathreplacing}

\newcommand{\sig}[0]{SIGMORPHON}
\newcommand{\tr}[0]{Transformer}
\newcommand{\onmt}[0]{OpenNMT}
\newcommand{\dataset}[0]{Czech OOV Inflection Dataset}
\newcommand{\morfoov}[0]{test-MorfFlex}
\newcommand{\realoov}[0]{test-neologisms}
\newcommand{\fullpar}[0]{full-paradigm}

\let\i\textit

\let\tag\texttt

\newcommand{\inflectlib}[0]{\footnote{\url{https://github.com/tomsouri/cz-inflect
}}}
\newcommand{\lindatdataset}[0]{\footnote{\url{http://hdl.handle.net/11234/1-5471}}}

\let\model\textsc
\newcommand{\copybs}[0]{\model{copy}}
\newcommand{\sklonuj}[0]{\model{sklonuj.cz}}
\newcommand{\nonneural}[0]{\model{SIG nonneur}}
\newcommand{\neural}[0]{\model{SIG trm}}
\newcommand{\enhanced}[0]{\model{SIG trm-tune}}
\newcommand{\retro}[0]{\model{Retro}}
\newcommand{\lstm}[0]{\model{LSTM}}
\newcommand{\trmmodel}[0]{\model{TRM}}

\newcommand{\ensemble}[0]{\model{Ensemble}}

\newcommand{\ssbetter}[0]{statistically significantly better}
\newcommand{\statsig}[0]{statistically significant}

\usepackage[inline]{enumitem}

\usepackage{booktabs}

\usepackage{calc}

\usepackage{amssymb}

\usepackage{cleveref}
\creflabelformat{equation}{#2\textup{#1}#3}

%% file: 00_abstract_short.tex
We focus on morphological inflection in out-of-vocabulary (OOV) conditions, an under-researched subtask in which state-of-the-art systems usually are less effective. We developed three systems: a retrograde model and two sequence-to-sequence (seq2seq) models based on LSTM and Transformer. For testing in OOV conditions, we automatically extracted a large dataset of nouns in the morphologically rich Czech language, with lemma-disjoint data splits, and we further manually annotated a real-world OOV dataset of neologisms.

In the standard OOV conditions, Transformer achieves the best results, with increasing performance in ensemble with LSTM, the retrograde model and SIGMORPHON baselines. On the real-world OOV dataset of neologisms, the retrograde model outperforms all neural models. Finally, our seq2seq models achieve state-of-the-art results in 9 out of 16 languages from SIGMORPHON 2022 shared task data in the OOV evaluation (feature overlap) in the large data condition.

We release the Czech OOV Inflection Dataset for rigorous evaluation in OOV conditions. Further, we release the inflection system with the seq2seq models as a ready-to-use Python library.

%% file: 01_intro.tex
\section{Introduction}
\label{sec:intro}

Inflection is a process of word formation in which a base word form (lemma) is modified to express grammatical categories. Many natural language generation systems that have natural text on the output, such as dialogue systems, need to be able to correctly inflect words. However, it has been shown that the state-of-the-art systems achieve rather poor results when tested on previously unseen inputs (OOV words) \citep{wug-test-liu-hulden-2022-transformer,goldman-etal-2022-un}. Despite an extensive exploration of the inflection task in recent years \citep{st16-cotterell-etal-2016-sigmorphon,st17-cotterell-etal-2017-conll,st18-cotterell-etal-2018-conll,st19-mccarthy-etal-2019-sigmorphon,st20-vylomova-etal-2020-sigmorphon,st21-pimentel-ryskina-etal-2021-sigmorphon,st22-kodner-etal-2022-sigmorphon,st23-goldman-etal-2023-sigmorphon} and outstanding results of the state-of-the-art systems, especially when the training data was plentiful \citep{wu-cotterell-transformer-2020,st21-pimentel-ryskina-etal-2021-sigmorphon}, the poor performance on OOV words has not been fully realized until recently, because the results had been inflated by the presence of training lemmas in the test dataset \citep{wug-test-liu-hulden-2022-transformer,goldman-etal-2022-un}.

To provide a consistent benchmark for inflection in OOV context, we release the \dataset{}\lindatdataset{} for rigorous evaluation, with a lemma-disjoint train-dev-test split of the pre-existing large morphological dictionary MorfFlex \citeplanguageresource{MorfFlexCZ20}. This benchmark is accompanied by a manually annotated small dataset of real-world OOV words (neologisms) in Czech. Unlike English, which has relatively simple morphology, e.g.\ just adding `-s' 
when forming the plural, the task of 
automatic inflection in morphologically rich languages such as Czech is quite difficult. An example of an inflection of a Czech neologism is shown in \cref{table:lingebra}.

\input{lingebra}

To our knowledge, this is the first dataset designed specifically for evaluation of inflection in the OOV conditions in Czech. 
In addition, Czech was not included in the 2022’s iteration of the SIGMORPHON shared task \citep{st22-kodner-etal-2022-sigmorphon}, which evaluated the performance of submitted systems on the implicit OOV subset of the shared task.

We develop three different systems, all data-driven, and compare them to several well-established systems, both Transformer-based and rule-based ones, in OOV conditions. We also build a state-of-the-art ready-to-use guesser for morphological inflection of Czech OOV nouns. 

Our first approach is a dictionary-based retrograde model: when given a lemma, search the database for a word that is most similar (has the longest common suffix), and inflect the lemma according to it.

The second and the third approach follow the standard neural approach using sequence-to-sequence architecture based on either LSTM \citep{lstm1997} or \tr{} \citep{vaswani2017attention}.

We adapt the systems to the\ OOV setting and tune them extensively. Then we evaluate them and compare them to one existing ready-to-use system, and to \sig{} shared task baselines \citep{st21-pimentel-ryskina-etal-2021-sigmorphon} on the \dataset{}.
Our systems either outperform the other evaluated systems or perform comparably.

In addition, we train and evaluate our neural setups on \sig{} 2022 shared task data (16 languages, Czech not included, all parts-of-speech) \citep{st22-kodner-etal-2022-sigmorphon} in the large training data condition, and in 9 languages we achieve state-of-the-art results in the OOV evaluation (feature overlap).

Finally, we address the lack of a reliable morpho-guesser for generation in Czech by releasing a ready-to-use Python library with our seq2seq models.\inflectlib{} We also release the \dataset{}.\lindatdataset{}

A more detailed description of our work, dataset creation, other exploratory experiments, as well as a profound summary of the related work, is provided in \citet{sourada23}.

We describe the new \dataset{} in \cref{sec:data} and our methodology in \cref{sec:methods}, with results and comparison of the three approaches in \cref{sec:comparison} and an error analysis in \cref{sec:error_analysis}. Finally, we conclude in \cref{sec:conclusion}.

%% file: lingebra.tex

\begin{table}
\centering
\begin{tabular}{l|ll}
\toprule
\multicolumn{3}{c}{LINGEBRA} \\
\midrule
Case / Number & Singular & Plural \\
\midrule
1.\ Nominative & lingebra & lingebry \\      
2.\ Genitive & lingebry & lingeber \\
3.\ Dative & lingebře & lingebrám \\
4.\ Accusative & lingebru &   lingebry   \\
5.\ Vocative & lingebro &   lingebry   \\
6.\ Locative & lingebře &   lingebrách   \\
7.\ Instrumental & lingebrou & lingebrami \\
\bottomrule
\end{tabular}

\caption{An example of an inflection of a Czech neologism not covered by the Czech inflection dictionary MorfFlex \protect\citeplanguageresource{MorfFlexCZ20}. ``\i{Lingebra}'' is a playful compound of words ''\i{lineární}" (linear) and ``\i{algebra}'' (algebra) (both in vocabulary; the neologism is inflected in the same way as ``\i{algebra}'').}
\label{table:lingebra}
\end{table}

%% file: 02_rel.tex
\section{Related Work}
\label{sec:related}

Earlier inflection systems were based on rules and dictionaries.
For Czech language, the linguistic module of the ASIMUT system \citep{asimut_kralikova_panevova_1990} determined inflection paradigms according to lemma endings based on the retrograde dictionary of \citet{retrogradni_slavickova_1975};
the sklonuj.cz system\footnote{\url{https://sklonuj.cz}} directly maps lemma endings to form endings based on hand-crafted rules.
Such simple ways of paradigm assignment have limited precision, often selecting an incorrect paradigm.
MorphoDiTa \citep{morphodita-strakova-2014-open}, on the other hand, only outputs inflections from the MorfFlex morphological dictionary \citeplanguageresource{MorfFlexCZ20}, leading to high precision but low recall, as no output is generated for OOV lemmas.
Later systems tried to extract and apply string transformation rules based on learned models \citep{flect-2013-robust,durrett-denero-2013-supervised}.

Since 2016, research has been considerably fueled by the annual \sig{} shared task on morphological inflection \citep{st16-cotterell-etal-2016-sigmorphon,st17-cotterell-etal-2017-conll,st18-cotterell-etal-2018-conll,st19-mccarthy-etal-2019-sigmorphon,st20-vylomova-etal-2020-sigmorphon,st21-pimentel-ryskina-etal-2021-sigmorphon,st22-kodner-etal-2022-sigmorphon,st23-goldman-etal-2023-sigmorphon}, including the release of datasets for up to 103 languages.
The increasingly prevalent approach has been the employment of the sequence-to-sequence (seq2seq) neural network architectures \citep{seq2seq}, often inspired by machine translation approaches, with hyperparameters adapted and tuned for the morphological inflection task.
The systems have been based on GRU or LSTM recurrent neural networks with attention \citep{faruqui2016morphological,st16-LMU-kann-schutze-2016-med}, and, since 2020 \citep{wu-cotterell-transformer-2020}, also the \tr{} \citep{vaswani2017attention}.
The models typically operate on sequences of individual characters and morphological labels, taking the lemma and the morphological information as input and producing the inflected form as the output.

Currently, the \tr{}-based systems seem to have almost completely mastered the task, achieving outstanding results especially when the training data is plentiful; with low training data and for unseen inputs (OOV words), the accuracies often plummet 
\citep{wug-test-liu-hulden-2022-transformer,goldman-etal-2022-un}. 
This can be partially alleviated by data augmentation techniques, such as data hallucination \citep{st22-UBC-yang-etal-2022-generalizing}, and by employing multilingual approaches. Since 2021, the \sig{} shared tasks include evaluation on unseen lemmas, but not for Czech language.

%% file: 03_data.tex
\section{\dataset{}}
\label{sec:data}
To allow for consistent evaluation of inflection in OOV conditions and for development of inflection systems, we create lemma-disjoint splits of
an existing morphological dictionary.
In addition, we annotate a small dataset of true OOV words (neologisms) to test the models in real-world conditions. The overview of the data splits is in \cref{tab:datasets}.
An example from the test-neologisms dataset can be found in \cref{tab:dataset_example}.

\input{datasets}
\input{dataset_example}

\subsection{MorfFlex Morphological Dictionary}

To build the train-dev-test split, we use the existing Czech morphological dictionary MorfFlex \citeplanguageresource{MorfFlexCZ20},
annotated with the morphological tagset of \citet{pos_hajic_2004}.
With more than 125M lemma-tag-form entries, it is relatively large compared to standard datasets in other languages.

We start by filtering the data by selecting the noun paradigm table entries (460k out of 1M entries). Of these, we removed all nonbasic-variant forms (such as nonstandard variants), all negation forms and malformed or deficient paradigm tables. The removed portion of the noun entries form 2\% of the noun entries and in the end, we acquired 449k noun paradigm tables. We then completed the incomplete paradigm tables (such as singular forms in tables of pluralia tantum or forms corresponding to non-flexible lemmas). 

We experimentally verified that omitting the negated variants of lemmas in training data does not have a negative impact on the performance of the models: we compared the performance of the inflection model (trained on data with no negations) on the standard development set and on a negated variant of it, and observed comparable results on both datasets.

We finish by randomly splitting the data into three lemma-disjoint parts: train, dev and test set with lemma counts in the ratio 8:1:1 (we denote the test set by \morfoov{} further in the text).

\subsection{Neologisms}

For evaluation in real-world OOV conditions, we build a new test set of true out-of-vocabulary words: neologisms. We considered several other options of what to use as the real OOV words, such as misspelled words, words with removed diacritics, proper nouns, but finally chose neologisms because they cannot be included in a dictionary by their very nature. 

We draw new words from a dictionary of Czech neologisms Čeština 2.0 \citep{HacknutaCestina}.\footnote{\url{https://cestina20.cz/}, in Czech only} Each entry contains the word or word phrase together with the explanation and usually also an example of usage in sentence or conversation.\footnote{For manual annotation, a subset of all neologisms, namely all words beginning with 'e' and 'j', was selected.} We randomly chose 101 lemmas corresponding to nouns (not word phrases) that are not present in MorfFlex \citeplanguageresource{MorfFlexCZ20}.

The inflected forms were first automatically generated by the rule-based guesser sklonuj.cz and then carefully post-edited by one annotator. The annotator was one of the authors, a senior undergraduate student and a Czech native speaker. The annotator was instructed to first post-edit the inflections and then revisit the annotations from a global perspective to ensure overall consistency of the inflections. In case of doubts, the annotator was encouraged to consult a standard reference of the Czech language, the Internet Language Reference Book,\footnote{\url{https://prirucka.ujc.cas.cz/en}} managed by Czech Language Institute of Czech Academy of Sciences. In case of multiple equally-correct forms in one paradigm cells, all of them were included.

By this process, we obtained the \realoov{} dataset. As it is disjoint from the training set and is drawn from a completely different source, it is expected to represent a greater challenge for the inflection systems. 

%% file: datasets.tex

\begin{table}
\centering
\setlength{\tabcolsep}{6pt}
\begin{tabular}{l r r r}
\toprule
 Set & lemmas & forms  & Source \\
\midrule
train & 360k & 5.04M  & MorfFlex \\
dev & 44k & 616k  & MorfFlex\\
test-MorfFlex & 44k & 616k & MorfFlex \\
test-\rlap{neologisms} & 101 & 1.4k  & Čeština 2.0  \\
\bottomrule
\end{tabular}
\caption{The four data splits, with lemma (paradigm table) counts and form counts.}
\label{tab:datasets}
\end{table}

%% file: dataset_example.tex

\begin{table}
\centering
\setlength{\tabcolsep}{6pt}
\begin{tabular}{l l l}
\toprule
 lemma & tag & form   \\
\midrule

elektrořidič & S1 & elektrořidič \\
elektrořidič & S2 & elektrořidiče \\
elektrořidič & S3 & elektrořidiči/elektrořidičovi \\
elektrořidič & S4 & elektrořidiče \\
elektrořidič & S5 & elektrořidiči \\
elektrořidič & S6 & elektrořidiči/elektrořidičovi \\
elektrořidič & S7 & elektrořidičem \\
elektrořidič & P1 & elektrořidiči/elektrořidičové \\
elektrořidič & P2 & elektrořidičů \\
elektrořidič & P3 & elektrořidičům \\
elektrořidič & P4 & elektrořidiče \\
elektrořidič & P5 & elektrořidiči/elektrořidičové \\
elektrořidič & P6 & elektrořidičích \\
elektrořidič & P7 & elektrořidiči \\

 \bottomrule
\end{tabular}
\caption{Example from the \realoov{} dataset. All 14 paradigm cells of lemma ``\i{elektrořidič}'' (driver of an electric car). The morphological tag is simplified to singular/plural (S/P) and 7 Czech cases (nominative, genitive, dative, accusative, vocative, locative and instrumental, numbered from 1 to 7). Some paradigm cells has two possible correct forms (those are separated by `/').}
\label{tab:dataset_example}
\end{table}

%% file: 04_methods.tex
\section{Methods}
\label{sec:methods}

\input{04a_eval_metrics}
\input{04b_baselines}
\input{04c_retro}
\input{04d_neural}
\input{04e_trm}
\input{04f_ensemble}

%% file: 04a_eval_metrics.tex
\subsection{Evaluation Metrics}
\label{sec:evaluation_metrics}

\paragraph{Form accuracy} (FA, see \cref{eq:fa}) is computed over all forms (except those marked as non-existent in the gold data).
A generated form is considered to be correct if it is equal to the gold form or if it is equal to one of the correct forms (in the case of the \realoov{} dataset which allows multiple gold forms in one paradigm cell).

\begin{equation}\label{eq:fa}
\textsc{FA} = \frac{\textsc{\#(correctly predicted forms)}}{\textsc{\#(all existent gold forms)}}
\end{equation}

\paragraph{Full-paradigm accuracy} (FPA, see \cref{eq:fpa}) is computed over all lemmas. A paradigm table generated for a lemma is considered to be correct if it contains correct form in every cell (except for the forms marked as non-existent in the gold data).

\begin{equation}\label{eq:fpa}
\textsc{FPA} = \frac{\textsc{\#(corr. predicted paradigm tables)}}{\textsc{\#(all lemmas)}}
\end{equation}

%% file: 04b_baselines.tex
\subsection{Baseline Systems}
\label{sec:baseline_systems}
We make use of several systems as baselines for performance comparison.

\paragraph{\copybs{}} The copy baseline ignores the training data and treats every lemma as inflexible during prediction: returns list of copies of the lemma as the predicted forms.

\paragraph{\sklonuj{}} Sklonuj.cz
represents the only ready-to-use guesser for Czech. It is based on hand-crafted rules and therefore has low recall. It does not use the training dataset.

\paragraph{\nonneural{}} The first standard baseline we use from \sig{} shared tasks \citep{st21-pimentel-ryskina-etal-2021-sigmorphon} is the non-neural one. 
It extracts transformation rules from the training examples and during prediction, it uses a majority classifier to apply the most frequent suitable rule.

\paragraph{\neural{}} Furthermore, we evaluate the neural baseline from \sig{} shared task \citep{st21-pimentel-ryskina-etal-2021-sigmorphon}, based on a vanilla Transformer with original hyperparameters from \citet{wu-cotterell-transformer-2021-applying}. 
The default training is with batch size 400 for 20k steps with the best performing checkpoint on the dev data, evaluated at the end of each epoch.

\paragraph{\enhanced{}} We observed that the default training setting is not ideal for our task, because on the training part of the \dataset{}, the optimizer finishes in less than 2 epochs. Consequently, we conducted experiments with increased batch size and the number of training steps and finally obtained best results with 150k train steps and batch size 800.

%% file: 04c_retro.tex
\subsection{Retrograde Approach}
\label{sec:retro}

The first approach finds a word in a database with the longest common suffix, and inflects according to it. We adapt the basic idea of the linguistic module in ASIMUT \citep{asimut_kralikova_panevova_1990}: deciding how the lemma inflects based on its ending segment. 
Unlike in ASIMUT, we do not extract the abstract paradigms manually but rather save all training words as possible paradigms and search amongst them for the most feasible during prediction. We call the approach Retrograde because it is based on retrograde lexicographical similarity of words and we denote the model \retro{}.

The model relies on two properties of Czech: (i) when two lemmas share the same ending, they also inflect identically, and (ii) during inflection (by number and case), only the ending changes while the rest of the word remains the same. This mostly holds in Czech but not in all other languages (e.g., semitic languages). The retrograde model is therefore strongly language dependent and we do not expect it to work well in all languages.

When building the model, we start with a morphological dictionary that contains complete paradigm tables for all covered lemmas. We save all the lemmas together with their inflection tables in a retrograde trie such that we can efficiently search them based on the suffixes.

When inflecting a lemma X, we search in the database for lemma A, such that X and A are most similar (have the longest common ending segment), and inflect lemma X according to the paradigm of lemma A. In case of multiple lemmas A in the dictionary with the same longest common ending segment with lemma X, we inflect X according to all of them and combine the predictions performing majority vote for each paradigm cell. In case of a tie, we choose the form from the most frequent ones randomly. 

The inflection of lemma X according to paradigm A is performed as follows (see \cref{tab:retro-inflection-by-paradigm}): remove the longest common suffix from lemma X and lemma A to obtain X-stem and A-stem. Then for each paradigm cell take the corresponding A-form and replace the A-stem by X-stem.

\input{retro}

\begin{figure}
  \includegraphics[width=1\hsize]{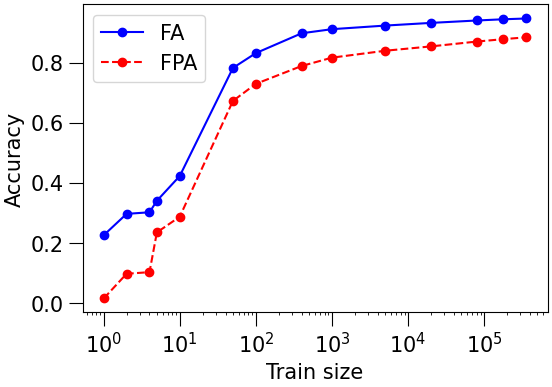}
  \caption[Retrograde model - dev evaluation]{Retrograde model development evaluation with different sizes of training dataset. Train size = count of paradigm tables in the training dataset.}
  \label{fig:retro-eval}
\end{figure}

We examined the dependence of the retrograde model on the size of the training data by experimenting with random subsets. As expected, the accuracy steadily improves when using more training data (\cref{fig:retro-eval}). 
Nevertheless, even with relatively small number of training lemmas (400 compared to the total 360k) the retrograde model outperforms the rule-based sklonuj.cz model.

%% file: retro.tex

\begin{table}[t]
\centering

\begin{tabular}{|ll|c|ll|}

\cline{1-2}\cline{4-5}
\multicolumn{2}{|c|}{\rule{0pt}{1em}HRAD}&  & \multicolumn{2}{c|}{\rule{0pt}{1em}ÚŘAD} \\ 

\cline{1-2}\cline{4-5}
\rule{0pt}{1em}hr-ad & \rule{0pt}{1em}hr-ady    &  & \rule{0pt}{1em}úř-ad & \rule{0pt}{1em}úř-ady \\
hr-adu &         &   & úř-adu & \\
hr-adu & ...     & $\longrightarrow$  & úř-adu & ...  \\
...  &          &  & ... & \\
hr-adem & hr-ady  &  & úř-adem & úř-ady \\
\cline{1-2}\cline{4-5}

\end{tabular}
\caption[Inflection according to a paradigm]{Retrograde model: example of inflection according to a paradigm. Lemma X for inflection: \textit{úřad} (office), found database lemma A with the longest common suffix: \textit{hrad} (castle), the longest common suffix: \textit{ad}.}
\label{tab:retro-inflection-by-paradigm}
\end{table}

%% file: 04d_neural.tex
\subsection{LSTM-Based seq2seq Models}
\label{sec:neural}

The second approach uses LSTM-based sequence-to-sequence (seq2seq) architectures originally proposed for the task of machine translation \citep{bahdanau-attn-2016neural}. These architectures were broadly used in the \sig{} shared tasks in recent years. We adapted the RNN-based encoder-decoder with soft attention as used by \citet{st16-LMU-kann-schutze-2016-med}. We used the implementation of the architectures as provided in the toolkit OpenNMT \citep{onmt-opennmt-klein-etal-2017}.\footnote{\texttt{OpenNMT-py v3.0.4}}

\input{mt}

\subsubsection{Source-Target Data Representation}
\label{sec:data-repre}

To be able to apply the MT architectures to our tasks, we formulate the inflection task as translation task using morphological tags, see \cref{fig:mt} for comparison of MT and inflection tasks, and for the example of input-output: the lemma plus tag as the source sequence, the inflected form as the target sequence.

Similarly to \citet{st16-LMU-kann-schutze-2016-med}, we use the individual characters of the source lemma followed by a separator and a 2-character morphological tag (describing the morphological categories of the target form) as input, and individual characters of the form as output. We investigated the usage of several different source-target representations, but obtained best results with this representation (although the differences in performance were marginal).

\subsubsection{Hyperparameters}

We perform hyperparameter tuning to adapt the architecture to the specifics of our task and the dataset. Batch size seems to be the most important: increasing it little by little from the original 20 to final 256 led to notable improvement, while adding the epochs by inflating the number of training steps with batch size fixed to 20 did not.

\citet{st16-LMU-kann-schutze-2016-med} used 1 layer with 100 GRU units both in the encoder (bi-directional) and the decoder.
We use LSTM units \citep{lstm1997} instead of GRU since it has been shown that LSTM performs better than GRU on larger datasets with shorter sequences \citep{lstm-gru-comparison}. 

Since our training dataset is much larger than the \sig{}'s 2016 dataset used by \citet{st16-LMU-kann-schutze-2016-med}, we examine extending the capacity of the network by increasing the number of hidden layers and their size and we experiment with the size of character and tag embeddings.
Since the input and the output sequence share most of the vocabulary, we experiment with shared embeddings.

We obtained the best result with LSTM-based seq2seq trained for 13 epochs with Adam \citep{Adam_Kingma_and_Ba_2015} with default values of $\beta$s, with learning rate 0.001 and warm-up 4k steps, batch 256, with 2 layers of size 200, shared embedding of dimension 128, bi-directional encoder and with Luong attention \citep{luong-attn-2015effective}; full configuration files are in the attachment. 

We denote this model \lstm{} further in the text.

%% file: mt.tex

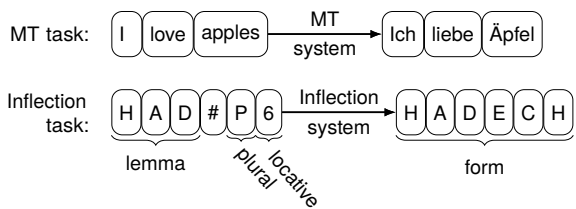
\begin{figure}
\begin{center}
\resizebox{1\hsize}{!}{%
\tikzstyle{rec}=[rectangle, rounded corners=1ex, minimum height = 0.8cm,anchor=center]
\begin{tikzpicture}[thin, inner sep=0.7ex]

\node[rec, text width = 1.4cm, align=right] (mt) {MT task:};
\node[rec, draw, right=0.3cm of mt, text width = 0.27cm] (i) {I};
\node[rec, draw, right=0cm of i] (love) {love};
\node[rec, draw, right=0cm of love] (apples) {apples};

\node[rec, draw, right=2cm of apples] (ich) {Ich};
\node[rec, draw, right=0cm of ich] (liebe) {liebe};
\node[rec, draw, right=0cm of liebe] (apfel) {Äpfel};

\draw [thick, -{Latex[length=2mm]}] (apples) -- (ich) node[midway, above] {MT};
\draw [thick, -{Latex[length=2mm]}] (apples) -- (ich) node[midway, below] {system};

\node[rec, below=0.5cm of mt, text width = 1.4cm, align=right] (T) {Inflection task:};
\node[rec, draw, right=0.3cm of T] (a) {H};
\node[rec, draw, right=0cm of a] (b) {A};
\node[rec, draw, right=0cm of b] (c) {D};
\node[rec, draw, right=0cm of c] (d) {\#};
\node[rec, draw, right=0cm of d] (e) {P};
\node[rec, draw, right=0cm of e] (f) {6};

\node[rec, draw, right=2cm of f] (g) {H};
\node[rec, draw, right=0cm of g] (h) {A};
\node[rec, draw, right=0cm of h] (i) {D};
\node[rec, draw, right=0cm of i] (j) {E};
\node[rec, draw, right=0cm of j] (k) {C};
\node[rec, draw, right=0cm of k] (l) {H};

\draw [thick, -{Latex[length=2mm]}] (f) -- (g) node[midway, above] {Inflection};
\draw [thick, -{Latex[length=2mm]}] (f) -- (g) node[midway, below] (arr) {system};

\draw[decorate,decoration={brace,amplitude=5pt, mirror}] (a.south west) -- (c.south east) node[midway, below=6pt] {lemma};

\draw[decorate,decoration={brace,amplitude=3pt, mirror}] (e.south west) -- (e.south east) node[midway, below=8pt] {};

\node[rectangle, rounded corners=1ex, rotate=310, below=0.4cm of f] (plural){plural};

\node[rectangle, rounded corners=1ex, rotate=310, above=0.2 cm of plural] (xxx){};
\node[rectangle, rounded corners=1ex, rotate=310, right=-0.2cm of xxx] {locative};

\draw[decorate,decoration={brace,amplitude=3pt, mirror}] (f.south west) -- (f.south east) node[midway, below=8pt] {};

\draw[decorate,decoration={brace,amplitude=5pt, mirror}] (g.south west) -- (l.south east) node[midway, below=8pt] {form};

\end{tikzpicture}
}
\end{center}
\caption{Input-output example for seq2seq models, and comparison of inflection task and the MT task. \textit{had} (snake) is the lemma, ``\#'' is a separator, \texttt{P6} is the morphological tag describing the target form (plural, locative case), \textit{hadech} (snakes used as in \textit{about snakes}) is the target inflected form.} 
\label{fig:mt}
\end{figure}

%% file: 04e_trm.tex
\subsection{Transformers}
We performed several experiments with current state-of-the-art Transformer-based seq2seq architecture implemented by \onmt{}. We used the same source-target data representation as for LSTM-based seq2seq models, described in \cref{sec:data-repre} (see \cref{fig:mt}).

Although \citet{wu-cotterell-transformer-2021-applying} claimed that a small-capacity \tr{} needs to be used in the inflection task, we achieved surprisingly good results with a high-capacity setting recommended for MT.\footnote{adapted from \url{https://github.com/ymoslem/OpenNMT-Tutorial/blob/main/2-NMT-Training.ipynb}} Only minor changes in the hyperparameters (hidden layer size, embedding dimension, dropouts, number of training steps and batch size) led to a model surpassing our extensively tuned \lstm{} model in both the form accuracy and the \fullpar{} accuracy.

The Transformer has the following parameters: 6 layers of size 256, trainable embeddings dimension 256 (for single-character tokens representing words and morphological tags),
8 attention heads, feed-forward of size 2048, trained for 40k steps with batch size 1024 and accumulation count 4 (effective batch size 4096) with Adam with ``noam'' decay, starting at learning rate 2, with $\beta_2=0.998$. For regularization it uses layer normalization and dropout 0.2, attention dropout 0.2 and label smoothing 0.1.
Full configuration files are in the attachment.

We denote this model \trmmodel{} further in the text.

%% file: 04f_ensemble.tex
\subsection{Model Ensembling}

In addition to experiments with individual models, we investigated combining all the baselines and our models into ensembles: for every target form, the combined models vote, and in case of a tie a random form from the most frequent predictions is chosen. 

We explored all possible combinations of models, and achieved best performance on the development set with the combination of two baselines and our 3 models: \nonneural{}, \enhanced{}, \retro{}, \lstm{} and \trmmodel{}. We denote it by \ensemble{} further in the text.

%% file: 05_eval.tex
\section{Results and Systems Comparison}
\label{sec:comparison}

We evaluated all our systems, the baselines and the \ensemble{} on the \morfoov{} dataset and the \realoov{} dataset, and compared the performance on both form accuracy (FA) and \fullpar{} accuracy (FPA). We measured statistical significance of the the differences on both metrics using the non-parametric approximation permutation test algorithm \citep{permutation-test, permutation-test-Gandy} with significance level 0.05 and with 10k resamplings. The results are presented in \cref{tab:eval-all}.

\input{eval-all}

\subsection{Test-MorfFlex}

On \morfoov{}, the best performing model is \trmmodel, which achieves 90.44\% in the \fullpar{} accuracy and is \ssbetter{} than all other models. In the form accuracy, it achieves 96.18\%, but \lstm{} and \enhanced{} perform only slightly worse, and the differences between them are not \statsig{}. All these three models are significantly better than the rest of the models. They are followed by \neural{} baseline and then by \retro{} model and \nonneural{} baseline. The \retro{} model statistically significantly outperforms the \nonneural{} baseline in both metrics. All models (except for the \copybs{} baseline) are significantly better than \sklonuj{}.

The success of the \tr{}-based models suggests that the \tr{} architecture is indeed suitable for the inflection task, even in the OOV conditions, at least when the training data is plentiful.

The \ensemble{} outperforms all the models in both metrics, showing that the errors made by the models are somehow complementary. Moreover, if we knew how to choose the best among the forms predicted by all the models and baselines, we could achieve 99.3\% in the form accuracy and 97.3\% in the \fullpar{} accuracy.


\subsection{Test-Neologisms}
The results are quire different on the \realoov{} dataset. There is a large drop in the performance of all models as compared to the performance on \morfoov{}, most pronounced in the performance of the neural models, especially \enhanced{} with almost 10\% drop in the form accuracy and 35\% in the \fullpar{} accuracy.

The \retro{} model and \nonneural{} baseline perform comparably and are \ssbetter{} in the form accuracy than all other models.\footnote{In the \fullpar{} accuracy, almost no difference was stastistically significant due to low number of paradigm tables in \realoov{} dataset.} 
The differences between the neural models and \sklonuj{} are not \statsig{}.

The overall drop in performance is understandable: the models were trained on data that come from the same distribution as \morfoov{}, but from a completely different distribution than \realoov{}.

The dominance of the \retro{} model and the \nonneural{} baseline could be (at least partially) caused by the fact that \realoov{} contains high percentage (37\%) of compounds, blends or words derived by prefixing, whose ending segment is an existing word present in MorfFlex (and thus possibly present in the training data). Those words are especially convenient for the \retro{} model since the simple algorithm is able to ignore the prefix and inflect the word correctly.

The \ensemble{} outperforms all models in the form accuracy, but not in the \fullpar{} accuracy. The upper bound accuracy, when choosing the predictions from all the models and baselines, is 96.5\% in the form accuracy and 82.2\% in the \fullpar{} accuracy, which shows that there is still room for improvement when using ensembles of current models.


\subsection{\sig{} 2022 Evaluation}

\input{05a_sig22eval}

%% file: eval-all.tex

\begin{table}
\centering
\begin{tabular}{lrrrr}
\toprule
test & \multicolumn{2}{c}{MorfFlex} & \multicolumn{2}{c}{neologisms} \\
\cmidrule(lr){2-3}\cmidrule(lr){4-5}
model & FA & FPA & FA & FPA \\
\midrule
\copybs{}   &   22.59   &  1.48                   & 13.13     &   0 \\

\sklonuj{}   &   88.88   & 74.43                  &  86.22     &   55 \\
\nonneural{}  &  94.78   &   88.15               & \textbf{89.49}  &   \textbf{71} \\
\neural{}   &   95.47   & 87.29               &   87.53   &   63  \\
\enhanced{} &   96.17   & 90.15             & 86.51    &   55  \\
\midrule
\retro{}  &   94.85   & 88.64    & 89.34     &   \textbf{71} \\
\lstm{}  &  96.16   &  89.80                 &  86.95     &   58 \\
\trmmodel{} & \textbf{96.18} & \textbf{90.44} &  87.24   &   61 \\
\midrule
\ensemble{} &  96.35 & 90.70   & 90.43 & 64 \\
\bottomrule
\end{tabular}
\caption{Evaluation of systems (FA and FPA in percent) on \morfoov{} (left) and \realoov{} (right). Upper section: SIG baseline systems by other authors \citep{st17-cotterell-etal-2017-conll,st21-pimentel-ryskina-etal-2021-sigmorphon}, trained on \dataset{}. Midsection: our systems. Bottom section: \ensemble{} combines \nonneural{}, \enhanced{}, \retro{}, \lstm{} and \trmmodel{}.
}
\label{tab:eval-all}
\end{table}

%% file: 05a_sig22eval.tex
\input{sig22_oov}
In order to evaluate the robustness of our seq2seq systems and to compare them to established approaches on a well-known dataset, we evaluated our \lstm{} and \trmmodel{} models on the \sig{} 2022 data \citep{st22-kodner-etal-2022-sigmorphon}. Specifically, we evaluate the performance on all 16 development languages\footnote{ang = Old English, ara = Modern Standard Arabic, asm = Assamese, got = Gothic, hun = Hungarian, kat = Georgian, khk = Khalkha Mongolian, kor = Korean, krl = Karelian, lud = Ludic, non = Old Norse, pol = Polish, poma = Pomak, slk = Slovak, tur = Turkish, vep = Veps \citep{st22-kodner-etal-2022-sigmorphon}} that included large training dataset and test data for the feature overlap (OOV) evaluation condition.

The datasets differ from our setting in several aspects: \begin{enumerate*}[label=(\roman*), itemjoin={{, }}, itemjoin*={{, and }}]
    \item the training dataset is smaller (\textasciitilde{}2k lemma-tag-form entries) compared to our dataset (\textasciitilde{}5M entries), even in the large data condition
    \item there are datasets for 16 different languages and none of them is Czech
    \item the data consist not only of nouns but also contain other \mbox{part-of-speech}
\end{enumerate*}.

To be able to run our models on the \sig{} data, we convert the data to our format by tokenizing the lemma and the word form to individual characters, add the special separator token to the end of the source sequence and then add the morphological features one by one. 
Once the model produces the output in our format, we convert it back to the \sig{} format and evaluate it using the official evaluation script.\footnote{\url{https://github.com/sigmorphon/2022InflectionST/tree/main/evaluation}}

We trained the \lstm{} model for 260k steps with batch size 256 (approx. 9.5k epochs), and the \trmmodel{} for 40k steps with effective batch size 4096 (approx. 23.4k epochs) and we chose the checkpoint with best performance on dev.

We present the results in the feature overlap (OOV) condition in \cref{tab:sig22-comp-feature}. We compare the performance of our systems with the neural and non-neural baseline and with all 5 submitted systems evaluated in the feature overlap (OOV) condition.

The \lstm{} model achieves the best score in 4 out of 16 languages and the \trmmodel{} model in other 5 languages. Averaged over all languages, our systems take the second and third place (\trmmodel{} with 86.1\%, \lstm{} with 85.3\%, respectively).

We suspect that the \tr{} approach lag behind LSTM in some languages might result from an interplay between the corpus size and the morphological complexity of the language. Some of the \sig{} corpora are relatively small for ML training. We hypothesize that \tr{}s might benefit from plentiful training data, but the influence of morphological complexity of the language remains to be accounted for.

It is also interesting that we achieved high score particularly in Slavic languages (Polish (pol), Pomak (poma) and Slovak (slk)). We can see that although we focused specifically on Czech morphology when tuning our setup, the models perform particularly well when trained and evaluated also on other Slavic languages.

These results show robustness of our seq2seq systems: although they were tuned for good performance on inflection of Czech nouns, they are suitable for inflecting also other parts-of-speech and other languages.

%% file: sig22_oov.tex

\begin{table*}
    
\centering

\begin{tabular}{c|ccccc|cc|cc} 
\toprule
& \multicolumn{5}{c|}{Submitted systems} & \multicolumn{2}{c|}{Baselines} & \multicolumn{2}{c}{Ours}\\
Lang & CLUZH & Flexica & OSU & TüM & UBC & Neural & NonNeur & \lstm{} & \trmmodel{}  \\
\midrule
ang   & \textbf{76.6} &   64.4   &   73.7   &   71.9   &   74.1   &   73.4   &   68.7   &  76.3  & 75.5 \\
ara   &   81.7   &   65.5   &   78.7   &   78.5   &   65.5   & 81.9 &   50.8   &  79.2  & \textbf{82.6} \\
asm   &   83.3   &   75.0   &   75.0   & \textbf{91.7} &   83.3   &   83.3   &   83.3   &  83.3  & 83.3 \\
got   &   92.9   &   41.4   & \textbf{94.1} &   91.7   &   91.7   &   93.5   &   87.6   &  92.3  & 92.3 \\
hun   &   93.5   &   62.9   &   93.1   &   92.8   &   91.5   &   \textbf{94.4}   &   73.1        & 92.8  & \textbf{94.4} \\
kat   &   96.7   &   95.7   &   96.7   &   96.7   &   96.7   & 97.3 &   96.7   & 97.3  & \textbf{97.8} \\
khk   & 94.1 & 47.1 & 94.1 & 94.1 & 88.2 & 94.1 & 88.2                                  & \textbf{100.0}  & 94.1 \\
kor   & \textbf{71.1} &   55.4   &   50.6   &   56.6   &   60.2   &   62.7   &   59.0   &  49.4  & 62.7 \\
krl   &   87.5 &   69.8   &   85.9   &   57.8   &   85.4   &   57.8   &   20.8          & \textbf{89.1}  & 85.9 \\
lud   &   87.3   &   92.0   &   92.9   &   93.4   &   88.2   & \textbf{94.3} &   93.4   &   89.2  & 92.0 \\
non   &   85.2   &   77.0   &   85.2   &   80.3   & \textbf{90.2} &   88.5   &   80.3   &   83.6  & 88.5 \\
pol   &   \textbf{96.1}   &   85.9   &   94.9   &   74.0   &   95.7   &   74.4   &   86.3         & \textbf{96.1}  & 95.6 \\
poma   & 76.1 &   54.5   &   70.1   &   69.4   &   73.3   &   74.1   &   47.8   & 75.2  & \textbf{76.3} \\
slk   &   93.5   &   90.0   &   92.2   &   70.4   & \textbf{95.7} &   71.1   &   92.4   &   95.2  & \textbf{95.7} \\
tur   &   93.7   &   57.9   & \textbf{95.2} &   80.2   &   92.9   &   79.4   &   66.7   & \textbf{95.2}  & 92.9 \\
vep   & \textbf{71.5} &   58.8   &   70.0   &   57.5   &   68.8   &   59.2   &   60.4   &  70.7  & 68.8 \\

\midrule
average & \textbf{86.3} & 68.3 & 83.9 & 78.6 & 83.8 & 80.0 & 72.2 & 85.3 & 86.1 \\
\bottomrule
\end{tabular}
\caption{\sig{} 2022 comparison -- Feature Overlap (FA metric in percent): A test pair’s feature set is attested in training, but its lemma is novel. Except for results of our systems, the table was adopted from \citet[Tables 17, 18 -- feature rows]{st22-kodner-etal-2022-sigmorphon}. The systems are: CLUZH \citep{wehrli-etal-2022-cluzh}, Flexica \citep{sherbakov-vylomova-2022-flexica}, OSU \citep{elsner-court-2022-osu}, TüM \citep{merzhevich-etal-2022-tum}, UBC \citep{st22-UBC-yang-etal-2022-generalizing}, and the \sig{} 2022 baselines \citep{st22-kodner-etal-2022-sigmorphon}.}
\label{tab:sig22-comp-feature}
\end{table*}

%% file: 06_error.tex
\section{Error Analysis}
\label{sec:error_analysis}

We perform error analysis of the model predictions on the dev set of the \dataset{}.

\subsection{Proper vs. Common Nouns}

Across all the models (except for \copybs{} baseline), almost 70\% or more of the incorrectly predicted forms are forms of proper nouns, while the total percentage of proper nouns in the dev set is only 31.68\%.

We compare the performance when evaluated on the corresponding parts of the dev set separately. The performance of all models improves substantially when running on common nouns only, and gets worse on the proper nouns subset. We show the differences of performance of the \trmmodel{} model in \cref{tab:proper-common}. Most noticable is the poor performance on the FPA on proper nouns. This trend is similar in the rest of the models, with the only exception of \sklonuj{}, which has extremely poor performance on proper nouns (72.30\% FA,  29.87\% FPA), but on common nouns it is much closer to the rest of the models (96.36\% FA,  87.40\% FPA). This is caused by the fact that it is not able to inflect a lot of proper nouns and simply returns nothing for them.

\input{proper-common}

\subsection{Distribution of Error Counts}

We focused on the error counts amongst the incorrectly generated paradigm tables, and on the percentages of errors made by the models in the individual paradigm cells (\tag{S1} up to \tag{P7}, \tag{S}=singular, \tag{P}=plural, \tag{1}-\tag{7}=case as in \cref{table:lingebra}).

The cells \tag{S6}, \tag{P1} and \tag{P7} are the most difficult to predict for all systems, while the easiest one is \tag{S1} (typically equal to the lemma).

\begin{figure}
  \centering
  \includegraphics[width=1\hsize]{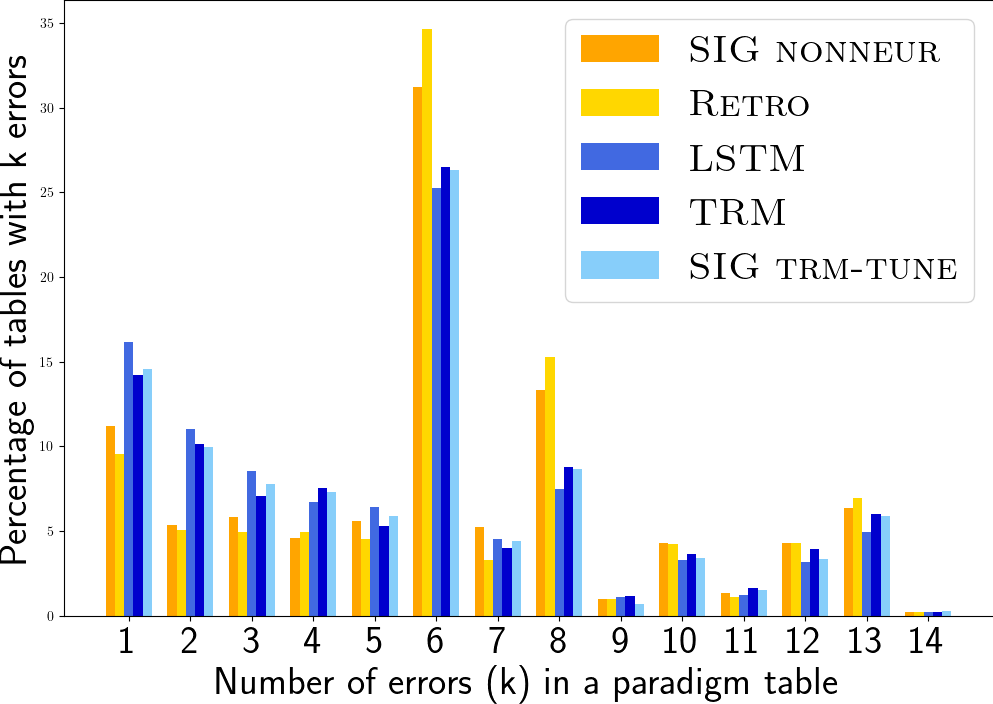}
  \caption{Histogram: the percentage of paradigm tables with given number of errors, counted for all incorrectly generated paradigm tables, for each model separately.}
  \label{fig:hist}
\end{figure}

We count the number of errors in each incorrectly predicted paradigm table, and for every error count (1 up to 14) we plot the percentage of tables with that number of errors for each of the models (\trmmodel{}, \enhanced{}, \lstm{}, \retro{} and \nonneural{}) separately (\cref{fig:hist}). Clearly, the most common number of errors amongst all models is 6 (more than 1/4 of all incorrect paradigm tables). More interestingly, the errors are in the same cells across all the models: for every model, more than 90\% of tables with 6 errors have the errors in the cells \tag{S2}, \tag{S3}, \tag{S4}, \tag{S6}, \tag{P1} and \tag{P5}. This probably reflects a property of the language itself: individual paradigms differ in these cells more than in the other cells.

Moreover, we can see that the non-neural models (\retro{}, \nonneural{}; shades of orange) behave similarly, the neural models (\lstm{}, \trmmodel{}, \enhanced{}; shades of blue) also behave similarly, but these two model groups behave differently. 
The neural models have higher percentage of small number of errors (1 up to 4 errors), while the non-neural tend to make more errors (especially 6, 8, and 13 errors). 
We believe this is because the neural models generate each form for a lemma independently without using the concept of paradigms, thus easily making occasional errors in individual forms. On the other hand, the non-neural models implicitly or explicitly use the concept of paradigms, and thus are more likely to either choose the paradigm correctly and make no errors, or incorrectly and make many errors.

%% file: proper-common.tex

\begin{table}
\centering
\begin{tabular}{lrr} 
\toprule
subset of dev set & FA & FPA \\
\midrule
proper nouns & 90.83 & 73.29 \\
common nouns & 98.40 & 93.91 \\
overall & 96.00 & 90.12 \\
\bottomrule
\end{tabular}
\caption{\trmmodel{} model performance on proper vs. common nouns.}
\label{tab:proper-common}
\end{table}

%% file: 07_concl.tex
\section{Conclusion}
\label{sec:conclusion}

We examined the understudied topic of inflection in out-of-vocabulary (OOV) conditions.

To this end, we created a lemma-disjoint train-dev-test split of a large pre-existing Czech morphological dictionary MorfFlex, and we also manually annotated a new small Czech test set of neologisms. We release this data as the \dataset{}.\lindatdataset{}

We studied three approaches to inflect OOVs: retrograde approach, LSTMs and Transformers. We thoroughly tested these approaches on our dataset, as well as OOV test sets for 16 other languages from the SIGMORPHON 2022 shared task.

We find that on our dataset, Transformer reaches the best results on \morfoov{}, whereas the retrograde approach beats both neural models on \realoov{}.
On the SIGMORPHON data, our seq2seq models achieve state-of-the-art results for 9 out of 16 languages.

We release our inflection system as a Python library.\inflectlib{}

%% file: 08_limitations.tex
\section*{Limitations}

As the \dataset{} encompasses all noun entries from the large Czech morphological dictionary MorfFlex \citeplanguageresource{MorfFlexCZ20}, with the exception of 2\% cleaned entries, we assume that we did not introduce any significant bias when constructing the dataset. 

The manually annotated \realoov{} is a subset of a corpus of Czech neologisms Čeština 2.0 \citep{HacknutaCestina}: for annotation, all words starting with 'e' and 'j' were selected. This process cannot be generally viewed as random and entirely representative. Nevertheless, we assume that the first character of a lemma does not have a significant influence on the way the word inflects. This assumption is supported by the fact that Czech is mostly a suffixing language. Another possible bias of the \realoov{} might be stemming from the fact that the underlying corpus of Czech neologisms contains many compounds.

Finally, the two most notable limitations of the \dataset{} is the restriction to nouns only, and the fact that it contains only the Czech language; we leave the other parts of speech and other languages for future work. We nevertheless assume that the presented results can be generalized to other languages, as evidenced by extensive evaluation of all methods also on 16 languages of the \sig{} shared task data.

Of the presented methods, the retrograde approach (\cref{sec:retro}) is expected to be the most limited in generalization across languages, as it exploits the shared similarity in suffix inflection between lemmas in the Czech language.

%% file: 09_ethical.tex
\section*{Ethical Considerations}

All manual annotations and evaluations within the work described in this paper were done by one male member of the team. However, as the morphological inflection in Czech is relatively straightforward and follows grammatical rules, we do not expect differences in annotation results in a mixed team.

No personal information has been among the lemmas extracted from the morphological dictionary. Both our neural methods, LSTM and Transformer, were trained from scratch on the training data, and we did not utilize any pre-trained LLMs, which might have contained personal information or biases.

Also, as we are not using any pre-trained LLMs, our methods are relatively cheap and efficient.

The authors declare that they are not aware of any conflict of interest related to the work published herein.

%% file: 10_ack.tex
\section*{Acknowledgements}

Computational resources for this work were provided by the e-INFRA CZ project (ID:90254), supported by the Ministry of Education, Youth and Sports of the Czech Republic.

The work described herein uses resources hosted by the
LINDAT/CLARIAH-CZ Research Infrastructure (projects LM2018101 and LM2023062, supported by the Ministry of Education, Youth and Sports of the Czech Republic).

This work has also been supported by the Grant Agency of the Czech Republic under the EXPRO program as project “LUSyD” (project No. GX20-16819X). 

We thank the Bernard Bolzano Endowment Fund for the contribution for covering the travel expenses to present this work.

We also thank the anonymous reviewers for their valuable comments.

%% file: main.bbl
\begin{thebibliography}{1}
\expandafter\ifx\csname natexlab\endcsname\relax\def\natexlab#1{#1}\fi

\bibitem[{Haji\v{c} et~al.(2020)Haji\v{c}, Hlavá\v{c}ov\'{a}, Mikulov\'{a}, Straka, and {\v{S}}t\v{e}p\'{a}nkov\'{a}}]{MorfFlexCZ20}
Haji\v{c}, Jan and Hlavá\v{c}ov\'{a}, Jaroslava and Mikulov\'{a}, Marie and Straka, Milan and {\v{S}}t\v{e}p\'{a}nkov\'{a}, Barbora. 2020.
\newblock \href {http://hdl.handle.net/11234/1-3185} {\emph{MorfFlex CZ 2.0}}.
\newblock Institute of Formal and Applied Linguistics, LINDAT/CLARIN, Charles University.

\end{thebibliography}


\begin{thebibliography}{37}
\expandafter\ifx\csname natexlab\endcsname\relax\def\natexlab#1{#1}\fi

\bibitem[{Bahdanau et~al.(2016)Bahdanau, Cho, and Bengio}]{bahdanau-attn-2016neural}
Dzmitry Bahdanau, Kyunghyun Cho, and Yoshua Bengio. 2016.
\newblock \href {http://arxiv.org/abs/1409.0473} {Neural machine translation by jointly learning to align and translate}.

\bibitem[{Cotterell et~al.(2018)Cotterell, Kirov, Sylak-Glassman, Walther, Vylomova, McCarthy, Kann, Mielke, Nicolai, Silfverberg, Yarowsky, Eisner, and Hulden}]{st18-cotterell-etal-2018-conll}
Ryan Cotterell, Christo Kirov, John Sylak-Glassman, G{\'e}raldine Walther, Ekaterina Vylomova, Arya~D. McCarthy, Katharina Kann, Sabrina~J. Mielke, Garrett Nicolai, Miikka Silfverberg, David Yarowsky, Jason Eisner, and Mans Hulden. 2018.
\newblock \href {https://doi.org/10.18653/v1/K18-3001} {The {C}o{NLL}{--}{SIGMORPHON} 2018 shared task: Universal morphological reinflection}.
\newblock In \emph{Proceedings of the {C}o{NLL}{--}{SIGMORPHON} 2018 Shared Task: Universal Morphological Reinflection}, pages 1--27, Brussels. Association for Computational Linguistics.

\bibitem[{Cotterell et~al.(2017)Cotterell, Kirov, Sylak-Glassman, Walther, Vylomova, Xia, Faruqui, K{\"u}bler, Yarowsky, Eisner, and Hulden}]{st17-cotterell-etal-2017-conll}
Ryan Cotterell, Christo Kirov, John Sylak-Glassman, G{\'e}raldine Walther, Ekaterina Vylomova, Patrick Xia, Manaal Faruqui, Sandra K{\"u}bler, David Yarowsky, Jason Eisner, and Mans Hulden. 2017.
\newblock \href {https://doi.org/10.18653/v1/K17-2001} {{C}o{NLL}-{SIGMORPHON} 2017 shared task: Universal morphological reinflection in 52 languages}.
\newblock In \emph{Proceedings of the {C}o{NLL} {SIGMORPHON} 2017 Shared Task: Universal Morphological Reinflection}, pages 1--30, Vancouver. Association for Computational Linguistics.

\bibitem[{Cotterell et~al.(2016)Cotterell, Kirov, Sylak-Glassman, Yarowsky, Eisner, and Hulden}]{st16-cotterell-etal-2016-sigmorphon}
Ryan Cotterell, Christo Kirov, John Sylak-Glassman, David Yarowsky, Jason Eisner, and Mans Hulden. 2016.
\newblock \href {https://doi.org/10.18653/v1/W16-2002} {The {SIGMORPHON} 2016 shared {T}ask{---}{M}orphological reinflection}.
\newblock In \emph{Proceedings of the 14th {SIGMORPHON} Workshop on Computational Research in Phonetics, Phonology, and Morphology}, pages 10--22, Berlin, Germany. Association for Computational Linguistics.

\bibitem[{Durrett and DeNero(2013)}]{durrett-denero-2013-supervised}
Greg Durrett and John DeNero. 2013.
\newblock \href {https://aclanthology.org/N13-1138} {Supervised learning of complete morphological paradigms}.
\newblock In \emph{Proceedings of the 2013 Conference of the North {A}merican Chapter of the Association for Computational Linguistics: Human Language Technologies}, pages 1185--1195, Atlanta, Georgia. Association for Computational Linguistics.

\bibitem[{Du{\v{s}}ek and Jur{\v{c}}{\'{i}}{\v{c}}ek(2013)}]{flect-2013-robust}
Ond{\v{r}}ej Du{\v{s}}ek and Filip Jur{\v{c}}{\'{i}}{\v{c}}ek. 2013.
\newblock \href {https://aclanthology.org/P13-3023} {Robust multilingual statistical morphological generation models}.
\newblock In \emph{51st Annual Meeting of the Association for Computational Linguistics Proceedings of the Student Research Workshop}, pages 158--164, Sofia, Bulgaria. Association for Computational Linguistics.

\bibitem[{Elsner and Court(2022)}]{elsner-court-2022-osu}
Micha Elsner and Sara Court. 2022.
\newblock \href {https://doi.org/10.18653/v1/2022.sigmorphon-1.22} {{OSU} at {S}ig{M}orphon 2022: Analogical inflection with rule features}.
\newblock In \emph{Proceedings of the 19th SIGMORPHON Workshop on Computational Research in Phonetics, Phonology, and Morphology}, pages 220--225, Seattle, Washington. Association for Computational Linguistics.

\bibitem[{Faruqui et~al.(2016)Faruqui, Tsvetkov, Neubig, and Dyer}]{faruqui2016morphological}
Manaal Faruqui, Yulia Tsvetkov, Graham Neubig, and Chris Dyer. 2016.
\newblock \href {http://arxiv.org/abs/1512.06110} {Morphological inflection generation using character sequence to sequence learning}.

\bibitem[{Fay and Follmann(2002)}]{permutation-test}
Michael~P. Fay and Dean~A. Follmann. 2002.
\newblock \href {http://www.jstor.org/stable/3087329} {Designing monte carlo implementations of permutation or bootstrap hypothesis tests}.
\newblock \emph{The American Statistician}, 56(1):63--70.

\bibitem[{Gandy(2009)}]{permutation-test-Gandy}
Axel Gandy. 2009.
\newblock \href {https://EconPapers.repec.org/RePEc:bes:jnlasa:v:104:i:488:y:2009:p:1504-1511} {Sequential implementation of monte carlo tests with uniformly bounded resampling risk}.
\newblock \emph{Journal of the American Statistical Association}, 104(488):1504--1511.

\bibitem[{Goldman et~al.(2023)Goldman, Batsuren, Khalifa, Arora, Nicolai, Tsarfaty, and Vylomova}]{st23-goldman-etal-2023-sigmorphon}
Omer Goldman, Khuyagbaatar Batsuren, Salam Khalifa, Aryaman Arora, Garrett Nicolai, Reut Tsarfaty, and Ekaterina Vylomova. 2023.
\newblock \href {https://doi.org/10.18653/v1/2023.sigmorphon-1.13} {{SIGMORPHON}{--}{U}ni{M}orph 2023 shared task 0: Typologically diverse morphological inflection}.
\newblock In \emph{Proceedings of the 20th SIGMORPHON workshop on Computational Research in Phonetics, Phonology, and Morphology}, pages 117--125, Toronto, Canada. Association for Computational Linguistics.

\bibitem[{Goldman et~al.(2022)Goldman, Guriel, and Tsarfaty}]{goldman-etal-2022-un}
Omer Goldman, David Guriel, and Reut Tsarfaty. 2022.
\newblock \href {https://doi.org/10.18653/v1/2022.acl-short.96} {(un)solving morphological inflection: Lemma overlap artificially inflates models{'} performance}.
\newblock In \emph{Proceedings of the 60th Annual Meeting of the Association for Computational Linguistics (Volume 2: Short Papers)}, pages 864--870, Dublin, Ireland. Association for Computational Linguistics.

\bibitem[{Haji\v{c}(2004)}]{pos_hajic_2004}
Jan Haji\v{c}. 2004.
\newblock \emph{Disambiguation of Rich Inflection (Computational Morphology of Czech)}.
\newblock Linguistic Data Consortium, University of Pennsylvania.

\bibitem[{Hochreiter and Schmidhuber(1997)}]{lstm1997}
Sepp Hochreiter and J\"{u}rgen Schmidhuber. 1997.
\newblock \href {https://doi.org/10.1162/neco.1997.9.8.1735} {Long short-term memory}.
\newblock \emph{Neural Comput.}, 9(8):1735–1780.

\bibitem[{Kann and Sch{\"u}tze(2016)}]{st16-LMU-kann-schutze-2016-med}
Katharina Kann and Hinrich Sch{\"u}tze. 2016.
\newblock \href {https://doi.org/10.18653/v1/W16-2010} {{MED}: The {LMU} system for the {SIGMORPHON} 2016 shared task on morphological reinflection}.
\newblock In \emph{Proceedings of the 14th {SIGMORPHON} Workshop on Computational Research in Phonetics, Phonology, and Morphology}, pages 62--70, Berlin, Germany. Association for Computational Linguistics.

\bibitem[{Kavka and Škrabal~et al.(2018)}]{HacknutaCestina}
Martin Kavka and Michal Škrabal~et al. 2018.
\newblock \emph{Hacknut\'{a} \v{c}e\v{s}tina}.
\newblock Jan Melvil Publishing.

\bibitem[{Kingma and Ba(2015)}]{Adam_Kingma_and_Ba_2015}
Diederik~P. Kingma and Jimmy Ba. 2015.
\newblock \href {http://arxiv.org/abs/1412.6980} {Adam: {A} method for stochastic optimization}.
\newblock In \emph{3rd International Conference on Learning Representations, {ICLR} 2015, San Diego, CA, USA, May 7-9, 2015, Conference Track Proceedings}.

\bibitem[{Klein et~al.(2017)Klein, Kim, Deng, Senellart, and Rush}]{onmt-opennmt-klein-etal-2017}
Guillaume Klein, Yoon Kim, Yuntian Deng, Jean Senellart, and Alexander Rush. 2017.
\newblock \href {https://www.aclweb.org/anthology/P17-4012} {{O}pen{NMT}: Open-source toolkit for neural machine translation}.
\newblock In \emph{Proceedings of {ACL} 2017, System Demonstrations}, pages 67--72, Vancouver, Canada. Association for Computational Linguistics.

\bibitem[{Kodner et~al.(2022)Kodner, Khalifa, Batsuren, Dolatian, Cotterell, Akkus, Anastasopoulos, Andrushko, Arora, Atanalov, Bella, Budianskaya, Ghanggo~Ate, Goldman, Guriel, Guriel, Guriel-Agiashvili, Kiera{\'s}, Krizhanovsky, Krizhanovsky, Marchenko, Markowska, Mashkovtseva, Nepomniashchaya, Rodionova, Scheifer, Sorova, Yemelina, Young, and Vylomova}]{st22-kodner-etal-2022-sigmorphon}
Jordan Kodner, Salam Khalifa, Khuyagbaatar Batsuren, Hossep Dolatian, Ryan Cotterell, Faruk Akkus, Antonios Anastasopoulos, Taras Andrushko, Aryaman Arora, Nona Atanalov, G{\'a}bor Bella, Elena Budianskaya, Yustinus Ghanggo~Ate, Omer Goldman, David Guriel, Simon Guriel, Silvia Guriel-Agiashvili, Witold Kiera{\'s}, Andrew Krizhanovsky, Natalia Krizhanovsky, Igor Marchenko, Magdalena Markowska, Polina Mashkovtseva, Maria Nepomniashchaya, Daria Rodionova, Karina Scheifer, Alexandra Sorova, Anastasia Yemelina, Jeremiah Young, and Ekaterina Vylomova. 2022.
\newblock \href {https://doi.org/10.18653/v1/2022.sigmorphon-1.19} {{SIGMORPHON}{--}{U}ni{M}orph 2022 shared task 0: Generalization and typologically diverse morphological inflection}.
\newblock In \emph{Proceedings of the 19th SIGMORPHON Workshop on Computational Research in Phonetics, Phonology, and Morphology}, pages 176--203, Seattle, Washington. Association for Computational Linguistics.

\bibitem[{Kr\'{a}l\'{i}kov\'{a} and Panevov\'{a}(1990)}]{asimut_kralikova_panevova_1990}
Kv\v{e}toslava Kr\'{a}l\'{i}kov\'{a} and Jarmila Panevov\'{a}. 1990.
\newblock {ASIMUT} - a method for automatic information retrieval from full texts.
\newblock \emph{Explizite Beschreibung der Sprache und automatische Textbearbeitung}, XVII.

\bibitem[{Liu and Hulden(2021)}]{wug-test-liu-hulden-2022-transformer}
Ling Liu and Mans Hulden. 2021.
\newblock \href {http://arxiv.org/abs/2104.06483} {Can a transformer pass the wug test? tuning copying bias in neural morphological inflection models}.
\newblock \emph{CoRR}, abs/2104.06483.

\bibitem[{Luong et~al.(2015)Luong, Pham, and Manning}]{luong-attn-2015effective}
Minh-Thang Luong, Hieu Pham, and Christopher~D. Manning. 2015.
\newblock \href {http://arxiv.org/abs/1508.04025} {Effective approaches to attention-based neural machine translation}.

\bibitem[{McCarthy et~al.(2019)McCarthy, Vylomova, Wu, Malaviya, Wolf-Sonkin, Nicolai, Kirov, Silfverberg, Mielke, Heinz, Cotterell, and Hulden}]{st19-mccarthy-etal-2019-sigmorphon}
Arya~D. McCarthy, Ekaterina Vylomova, Shijie Wu, Chaitanya Malaviya, Lawrence Wolf-Sonkin, Garrett Nicolai, Christo Kirov, Miikka Silfverberg, Sabrina~J. Mielke, Jeffrey Heinz, Ryan Cotterell, and Mans Hulden. 2019.
\newblock \href {https://doi.org/10.18653/v1/W19-4226} {The {SIGMORPHON} 2019 shared task: Morphological analysis in context and cross-lingual transfer for inflection}.
\newblock In \emph{Proceedings of the 16th Workshop on Computational Research in Phonetics, Phonology, and Morphology}, pages 229--244, Florence, Italy. Association for Computational Linguistics.

\bibitem[{Merzhevich et~al.(2022)Merzhevich, Gbadegoye, Girrbach, Li, and Shim}]{merzhevich-etal-2022-tum}
Tatiana Merzhevich, Nkonye Gbadegoye, Leander Girrbach, Jingwen Li, and Ryan Soh-Eun Shim. 2022.
\newblock \href {https://doi.org/10.18653/v1/2022.sigmorphon-1.20} {{SIGMORPHON} 2022 task 0 submission description: Modelling morphological inflection with data-driven and rule-based approaches}.
\newblock In \emph{Proceedings of the 19th SIGMORPHON Workshop on Computational Research in Phonetics, Phonology, and Morphology}, pages 204--211, Seattle, Washington. Association for Computational Linguistics.

\bibitem[{Pimentel et~al.(2021)Pimentel, Ryskina, Mielke, Wu, Chodroff, Leonard, Nicolai, Ghanggo~Ate, Khalifa, Habash, El-Khaissi, Goldman, Gasser, Lane, Coler, Oncevay, Montoya~Samame, Silva~Villegas, Ek, Bernardy, Shcherbakov, Bayyr-ool, Sheifer, Ganieva, Plugaryov, Klyachko, Salehi, Krizhanovsky, Krizhanovsky, Vania, Ivanova, Salchak, Straughn, Liu, Washington, Ataman, Kiera{\'s}, Woli{\'n}ski, Suhardijanto, Stoehr, Nuriah, Ratan, Tyers, Ponti, Aiton, Hatcher, Prud{'}hommeaux, Kumar, Hulden, Barta, Lakatos, Szolnok, {\'A}cs, Raj, Yarowsky, Cotterell, Ambridge, and Vylomova}]{st21-pimentel-ryskina-etal-2021-sigmorphon}
Tiago Pimentel, Maria Ryskina, Sabrina~J. Mielke, Shijie Wu, Eleanor Chodroff, Brian Leonard, Garrett Nicolai, Yustinus Ghanggo~Ate, Salam Khalifa, Nizar Habash, Charbel El-Khaissi, Omer Goldman, Michael Gasser, William Lane, Matt Coler, Arturo Oncevay, Jaime~Rafael Montoya~Samame, Gema~Celeste Silva~Villegas, Adam Ek, Jean-Philippe Bernardy, Andrey Shcherbakov, Aziyana Bayyr-ool, Karina Sheifer, Sofya Ganieva, Matvey Plugaryov, Elena Klyachko, Ali Salehi, Andrew Krizhanovsky, Natalia Krizhanovsky, Clara Vania, Sardana Ivanova, Aelita Salchak, Christopher Straughn, Zoey Liu, Jonathan~North Washington, Duygu Ataman, Witold Kiera{\'s}, Marcin Woli{\'n}ski, Totok Suhardijanto, Niklas Stoehr, Zahroh Nuriah, Shyam Ratan, Francis~M. Tyers, Edoardo~M. Ponti, Grant Aiton, Richard~J. Hatcher, Emily Prud{'}hommeaux, Ritesh Kumar, Mans Hulden, Botond Barta, Dorina Lakatos, G{\'a}bor Szolnok, Judit {\'A}cs, Mohit Raj, David Yarowsky, Ryan Cotterell, Ben Ambridge, and Ekaterina Vylomova. 2021.
\newblock \href {https://doi.org/10.18653/v1/2021.sigmorphon-1.25} {{SIGMORPHON} 2021 shared task on morphological reinflection: Generalization across languages}.
\newblock In \emph{Proceedings of the 18th SIGMORPHON Workshop on Computational Research in Phonetics, Phonology, and Morphology}, pages 229--259, Online. Association for Computational Linguistics.

\bibitem[{Sherbakov and Vylomova(2022)}]{sherbakov-vylomova-2022-flexica}
Andreas Sherbakov and Ekaterina Vylomova. 2022.
\newblock \href {https://doi.org/10.18653/v1/2022.sigmorphon-1.25} {Morphology is not just a naive {B}ayes {--} {U}ni{M}elb submission to {SIGMORPHON} 2022 {ST} on morphological inflection}.
\newblock In \emph{Proceedings of the 19th SIGMORPHON Workshop on Computational Research in Phonetics, Phonology, and Morphology}, pages 240--246, Seattle, Washington. Association for Computational Linguistics.

\bibitem[{Slav\'{i}\v{c}kov\'{a}(1975)}]{retrogradni_slavickova_1975}
Eleonora Slav\'{i}\v{c}kov\'{a}. 1975.
\newblock \emph{Retrogr\'{a}dn\'{i} morfematick\'{y} slovn\'{i} \v{c}e\v{s}tiny}, 1. edition.
\newblock Academia.

\bibitem[{Sourada(2023)}]{sourada23}
Tom\'{a}\v{s} Sourada. 2023.
\newblock \href {http://hdl.handle.net/20.500.11956/184286} {\emph{Automatic inflection in {C}zech language}}.
\newblock Charles University, Faculty of Mathematics and Physics, Institute of Formal and Applied Linguistics, Prague.

\bibitem[{Strakov{\'a} et~al.(2014)Strakov{\'a}, Straka, and Haji{\v{c}}}]{morphodita-strakova-2014-open}
Jana Strakov{\'a}, Milan Straka, and Jan Haji{\v{c}}. 2014.
\newblock \href {https://doi.org/10.3115/v1/P14-5003} {Open-source tools for morphology, lemmatization, {POS} tagging and named entity recognition}.
\newblock In \emph{Proceedings of 52nd Annual Meeting of the Association for Computational Linguistics: System Demonstrations}, pages 13--18, Baltimore, Maryland. Association for Computational Linguistics.

\bibitem[{Sutskever et~al.(2014)Sutskever, Vinyals, and Le}]{seq2seq}
Ilya Sutskever, Oriol Vinyals, and Quoc~V. Le. 2014.
\newblock \href {http://arxiv.org/abs/1409.3215} {Sequence to sequence learning with neural networks}.

\bibitem[{Vaswani et~al.(2017)Vaswani, Shazeer, Parmar, Uszkoreit, Jones, Gomez, Kaiser, and Polosukhin}]{vaswani2017attention}
Ashish Vaswani, Noam Shazeer, Niki Parmar, Jakob Uszkoreit, Llion Jones, Aidan~N. Gomez, Lukasz Kaiser, and Illia Polosukhin. 2017.
\newblock \href {http://arxiv.org/abs/1706.03762} {Attention is all you need}.

\bibitem[{Vylomova et~al.(2020)Vylomova, White, Salesky, Mielke, Wu, Ponti, Maudslay, Zmigrod, Valvoda, Toldova, Tyers, Klyachko, Yegorov, Krizhanovsky, Czarnowska, Nikkarinen, Krizhanovsky, Pimentel, Torroba~Hennigen, Kirov, Nicolai, Williams, Anastasopoulos, Cruz, Chodroff, Cotterell, Silfverberg, and Hulden}]{st20-vylomova-etal-2020-sigmorphon}
Ekaterina Vylomova, Jennifer White, Elizabeth Salesky, Sabrina~J. Mielke, Shijie Wu, Edoardo~Maria Ponti, Rowan~Hall Maudslay, Ran Zmigrod, Josef Valvoda, Svetlana Toldova, Francis Tyers, Elena Klyachko, Ilya Yegorov, Natalia Krizhanovsky, Paula Czarnowska, Irene Nikkarinen, Andrew Krizhanovsky, Tiago Pimentel, Lucas Torroba~Hennigen, Christo Kirov, Garrett Nicolai, Adina Williams, Antonios Anastasopoulos, Hilaria Cruz, Eleanor Chodroff, Ryan Cotterell, Miikka Silfverberg, and Mans Hulden. 2020.
\newblock \href {https://doi.org/10.18653/v1/2020.sigmorphon-1.1} {{SIGMORPHON} 2020 shared task 0: Typologically diverse morphological inflection}.
\newblock In \emph{Proceedings of the 17th SIGMORPHON Workshop on Computational Research in Phonetics, Phonology, and Morphology}, pages 1--39, Online. Association for Computational Linguistics.

\bibitem[{Wehrli et~al.(2022)Wehrli, Clematide, and Makarov}]{wehrli-etal-2022-cluzh}
Silvan Wehrli, Simon Clematide, and Peter Makarov. 2022.
\newblock \href {https://doi.org/10.18653/v1/2022.sigmorphon-1.21} {{CLUZH} at {SIGMORPHON} 2022 shared tasks on morpheme segmentation and inflection generation}.
\newblock In \emph{Proceedings of the 19th SIGMORPHON Workshop on Computational Research in Phonetics, Phonology, and Morphology}, pages 212--219, Seattle, Washington. Association for Computational Linguistics.

\bibitem[{Wu et~al.(2020)Wu, Cotterell, and Hulden}]{wu-cotterell-transformer-2020}
Shijie Wu, Ryan Cotterell, and Mans Hulden. 2020.
\newblock \href {http://arxiv.org/abs/2005.10213} {Applying the transformer to character-level transduction}.
\newblock \emph{CoRR}, abs/2005.10213.

\bibitem[{Wu et~al.(2021)Wu, Cotterell, and Hulden}]{wu-cotterell-transformer-2021-applying}
Shijie Wu, Ryan Cotterell, and Mans Hulden. 2021.
\newblock \href {https://doi.org/10.18653/v1/2021.eacl-main.163} {Applying the transformer to character-level transduction}.
\newblock In \emph{Proceedings of the 16th Conference of the European Chapter of the Association for Computational Linguistics: Main Volume}, pages 1901--1907, Online. Association for Computational Linguistics.

\bibitem[{Yang et~al.(2022)Yang, Yang, Nicolai, and Silfverberg}]{st22-UBC-yang-etal-2022-generalizing}
Changbing Yang, Ruixin~(Ray) Yang, Garrett Nicolai, and Miikka Silfverberg. 2022.
\newblock \href {https://doi.org/10.18653/v1/2022.sigmorphon-1.23} {Generalizing morphological inflection systems to unseen lemmas}.
\newblock In \emph{Proceedings of the 19th SIGMORPHON Workshop on Computational Research in Phonetics, Phonology, and Morphology}, pages 226--235, Seattle, Washington. Association for Computational Linguistics.

\bibitem[{Yang et~al.(2020)Yang, Yu, and Zhou}]{lstm-gru-comparison}
Shudong Yang, Xueying Yu, and Ying Zhou. 2020.
\newblock \href {https://doi.org/10.1109/IWECAI50956.2020.00027} {Lstm and gru neural network performance comparison study: Taking yelp review dataset as an example}.
\newblock In \emph{2020 International Workshop on Electronic Communication and Artificial Intelligence {IWECAI}}, pages 98--101.

\end{thebibliography}
